\documentclass[12pt,a4paper]{article}
\usepackage{float}

\usepackage[T1]{fontenc}
\usepackage[utf8]{inputenc}

\usepackage{newtxtext}
\usepackage{newtxmath}

\usepackage[
a4paper,
left=2.54cm,
right=2.54cm,
top=2.54cm,
bottom=2.54cm
]{geometry}

\usepackage{setspace}
\usepackage{amsmath}
\usepackage{amssymb}

\usepackage{booktabs}
\usepackage{tabularx}
\usepackage{longtable}
\usepackage{array}

\usepackage{graphicx}
\usepackage{float}

\usepackage{enumitem}

\usepackage[hidelinks]{hyperref}

\usepackage[
backend=biber,
style=apa,
sorting=nyt
]{biblatex}

\DeclareLanguageMapping{english}{english-apa}
\newcolumntype{Y}{>{\raggedright\arraybackslash}X}

\begin{document}


\begin{center}

{\Large\bfseries
An Evidence-Grounded Retrieval-Augmented Transformer Framework for Health Misinformation Verification
}

\vspace{0.4cm}

$^{1,2}$ Isah M. Bukar,
$^{1,2,3}$ Bala Mairiga Abduljalil,
$^{1,2}$ Bashir Saleh Maina,
$^{4}$ Abdulbasit Hassan

\vspace{0.2cm}

{\small
$^{1}$ Department of Computer Science, Yobe State University, Damaturu, Nigeria\\
$^{2}$ Yobe AI Research Laboratory (YAIR Lab), Yobe State University, Damaturu, Nigeria\\
$^{3}$ Center for AI \& Big Data, Novosibirsk State University, Novosibirsk, Russian Federation\\
$^{4}$ College of Computer Science and Engineering, Hamad Bin Khalifa University, Doha, Qatar
}

\vspace{0.2cm}

\noindent
\textbf{Corresponding author:}\\
\texttt{bsmaina@ysu.edu.ng}

\vspace{0.3cm}

\end{center}

\begin{abstract}
\noindent
The rapid spread of false and misleading health information through digital platforms has become a major public health challenge, particularly during infectious disease outbreaks where delayed verification can influence public behaviour and hinder effective disease control. Although recent advances in automated health misinformation detection have shown encouraging results, most existing approaches rely heavily on global biomedical resources and often fail to capture the local context needed to verify claims in developing countries. This study presents a retrieval-augmented transformer framework designed to verify health-related claims using trusted evidence from the World Health Organization and the Nigeria Centre for Disease Control and Prevention. The framework combines semantic evidence retrieval with transformer-based classification to determine whether a claim is true, false, or misleading. To evaluate the proposed approach, a manually annotated dataset of 67 verified health claims covering coronavirus disease, Lassa fever, cholera, measles, and monkeypox was compiled from Nigerian fact-checking sources. Three transformer models and a retrieval-augmented configuration were evaluated. The Bidirectional Encoder Representations from Transformers model achieved the best performance, with an accuracy of 71\% and a weighted F1-score of 0.66. Although retrieval augmentation did not improve classification performance because the current evidence repository was limited in size and coverage, the findings highlight the importance of comprehensive and authoritative knowledge sources for reliable health misinformation verification. The proposed framework provides a practical foundation for developing context-aware and evidence-driven health misinformation verification systems for Nigeria and other resource-constrained settings.

\end{abstract}


\section{Introduction}

\subsection{Background and Context}

The World Health Organization defines an infodemic as an overabundance of information, both accurate and inaccurate, that makes it difficult for people to find trustworthy guidance during a health emergency. This challenge is not abstract in the Nigerian context. In April 2024, a video alleging a haemorrhagic fever outbreak had killed a student at the Federal Teaching Hospital in Lokoja, Kogi State, circulated widely on social media before the Kogi State Commissioner for Health issued an official clarification: the patient's Lassa fever test had returned negative, and no outbreak had occurred in the state \parencite{kogistate2024}. Similar patterns recur across other outbreak-prone diseases. During the 2022 mpox resurgence, claims circulated that the outbreak originated from United States-funded biological laboratories in Nigeria, a claim later shown to have no genetic or epidemiological support \parencite{factcheckhub2022}. During the 2024 cholera outbreak in Lagos State, tiger-nut drinks were widely blamed as the confirmed cause, a claim that oversimplified a more complex link to contaminated water used in preparation \parencite{africacheck2024}. Claims about COVID-19 vaccines, including that they alter human DNA, cause infertility, or contain tracking microchips, have circulated in Nigeria since 2020 despite repeated rebuttals from national health authorities \parencite{dubawa2022}.

These cases share a common structure: an initial claim spreads faster than verification, and the eventual correction depends on a government agency or journalist manually tracing the claim back to authoritative records, often days or weeks after the claim has already shaped public behaviour. Given that Nigeria carries one of the highest documented burdens of several of these diseases, including Lassa fever and measles, the cost of delayed or absent verification is measured not only in public trust but in health outcomes.

\subsection{Problem Statement and Research Gap}

Automated approaches to health misinformation verification have advanced considerably in recent years. Retrieval-augmented generation (RAG) systems have been shown to improve fact-checking accuracy by grounding large language model outputs in trusted document collections rather than relying on parametric knowledge alone \parencite{li2025rag}. Transformer-based classifiers have been applied to distinguish credible from non-credible online health content using sources such as PubMed \parencite{bayani2025transformer}, and benchmark datasets such as PUBHEALTH have enabled veracity prediction with accompanying explanation generation \parencite{kotonya2020explainable}. Graph-based and multi-evidence RAG variants have further reduced hallucination rates in public health question answering \parencite{xu2025megarag, hang2025trumorgpt}.

However, these systems are built almost exclusively around global, English-language biomedical literature repositories such as PubMed and Scopus. None are grounded in the combination of WHO guidance and a national public health authority's outbreak-specific reporting, and none have been evaluated against claims drawn from a Nigerian information environment. This leaves a gap: a verification framework that pairs global authoritative guidance (WHO) with localised, outbreak-specific evidence (NCDC situation reports), the two levels of evidence actually needed to assess claims such as the Kogi State Lassa fever case above, which required both general disease knowledge and location-specific outbreak data to resolve.

\subsection{Aim, Objectives and Research Questions}

This study aims to design and evaluate a retrieval-augmented transformer framework that verifies health-related claims against WHO and NCDC knowledge sources. The specific objectives are to:

\begin{enumerate}[label=(\roman*)]
    \item construct a structured knowledge base from WHO fact sheets and NCDC situation reports covering five outbreak-prone diseases (Lassa fever, cholera, measles, mpox, and COVID-19);
    \item compile a labelled claim dataset drawn from verified Nigerian fact-checking sources, spanning false, misleading, and true claims;
    \item implement a retrieval-augmented transformer that classifies claims against retrieved WHO/NCDC evidence; and
    \item evaluate the framework's classification performance against the labelled dataset.
\end{enumerate}

Accordingly, the study addresses the following research questions:

\begin{enumerate}[label=(\roman*)]
    \item How accurately can a retrieval-augmented transformer classify Nigerian health claims when grounded in WHO and NCDC sources?
    \item Does combining global (WHO) and local (NCDC) evidence improve verification accuracy compared to either source alone?
\end{enumerate}

\subsection{Significance and Scope of the Study}

This study contributes to the growing body of work on automated health misinformation verification by addressing a gap specific to low-resource, regionally-grounded public health communication. Its findings are intended to inform future tools that support NCDC risk communication efforts and Nigerian fact-checking organisations such as Dubawa and Africa Check, whose current verification processes remain manual. The scope of the study is bounded to five diseases and to claims sourced from established Nigerian fact-checking platforms; it does not attempt to cover all disease categories or all languages in which health misinformation circulates in Nigeria.

The remainder of this paper is organised as follows: Section 2 reviews related literature and theoretical frameworks underpinning retrieval-augmented verification systems; Section 3 describes the methodology, including knowledge base construction and model implementation; Section 4 presents results and discussion; and Section 5 concludes with recommendations and directions for future research.

\section{Literature Review and Theoretical Framework}

\subsection{Health Misinformation and Fact Verification}

Health misinformation detection has developed along several methodological lines beyond simple keyword or source-blacklisting approaches. \textcite{ulhussna2024dissecting} synthesise 87 studies on COVID-19 misinformation detection on X (formerly Twitter), consolidating machine learning and deep learning approaches, dataset characteristics, and disseminator behaviour into a combined training resource, while identifying persistent gaps in cross-platform generalisability. \textcite{ahmad2026misinformation} apply transformer-based pretrained language models, including BERT, GPT-2, and XLNet, to detect and classify misinformation in the context of the 2020 United States election, further demonstrating that a combined moderation intervention could reduce misinformation prevalence by 87.9\% within a 40-minute intervention window.

Affective and behavioural signals have also proven useful beyond purely textual features. \textcite{liu2024emotion} review the growing body of work that incorporates emotional and sentiment signals into misinformation detection, arguing that affective cues such as fear or outrage are consistently associated with the spread of false health content and that fusing these signals with textual features improves detection robustness. Collectively, this literature establishes that health misinformation detection benefits from combining linguistic, affective, and behavioural signals, though most systems are trained on general-purpose datasets rather than claims already verified against authoritative health institutions, which motivates the source-grounded approaches reviewed in the following subsections.

\subsection{Transformer Models for Health Misinformation Detection}

Transformer architectures have become the dominant modelling choice for health misinformation classification tasks specifically. \textcite{gundapu2021transformer} demonstrate this with an ensemble of BERT, ALBERT, and XLNet transformers fine-tuned for COVID-19 fake news detection, achieving an F1 score of 0.9855 and ranking fifth of 160 teams in the ConstraintAI 2021 shared task, illustrating the strong ceiling performance transformer ensembles can reach on well-curated benchmark data.

More directly relevant to the present study, \textcite{bayani2025transformer} take a source-grounded approach, using PubMed as a trusted reference corpus and applying BERT, BioBERT, and SciBERT classifiers alongside cosine and Jaccard similarity measures to automatically assess the credibility of online health web pages. This architecture directly anticipates the retrieval-grounded verification approach adopted in this study: pairing a transformer classifier with a trusted external source, rather than relying on the classifier's parametric knowledge alone. Together, these studies confirm that transformer-based classifiers, whether used alone or in ensembles, consistently outperform earlier machine learning baselines on health misinformation tasks, and that grounding classification in a trusted external source improves reliability further still.

\subsection{Retrieval-Augmented Generation in Knowledge-Intensive NLP}

Retrieval-augmented generation (RAG) has emerged as the principal architecture for grounding large language model outputs in verifiable external evidence rather than relying solely on parametric knowledge. \textcite{li2025rag} demonstrate this directly for health fact-checking, integrating a RAG pipeline with GPT-4 over approximately 130,000 peer-reviewed COVID-19 papers drawn from PubMed and Scopus, and comparing naive, LOTR-RAG, CRAG, and SRAG variants to show measurable gains in fact-checking accuracy. \textcite{hang2025trumorgpt} extend the retrieval component itself, proposing TrumorGPT, a graph-based RAG framework built over a continuously updated semantic health knowledge graph capable of verifying "trumors" (true rumours) as well as false ones, addressing a nuance that binary true/false classifiers typically miss. \textcite{xu2025megarag} further advance retrieval design with MEGA-RAG, which combines dense retrieval, keyword search, biomedical knowledge graphs, and cross-encoder reranking with a discrepancy-aware refinement module, reporting a reduction in large language model hallucinations of more than 40\% in public health question answering.

The broader applicability of RAG across healthcare sub-domains is documented in the systematic review by \textcite{aboelenen2025survey}, who categorise Naive, Advanced, and Modular RAG architectures across text, visual, and hybrid medical applications while highlighting persistent evaluation and source-interpretability challenges that the field has yet to resolve. Across this literature, a clear pattern emerges: RAG systems consistently outperform non-retrieval baselines on knowledge-intensive verification tasks, and multi-source or graph-augmented retrieval designs offer further gains over single-corpus retrieval. However, every system reviewed here retrieves from either global biomedical literature or general-purpose knowledge graphs; none retrieves from the combination of a global public health authority (WHO) and a national disease control agency (NCDC) that this study requires.

\subsection{Trusted Health Knowledge Sources: WHO and NCDC}

A recurring theme across the RAG and fact-verification literature is that system reliability depends heavily on the trustworthiness and provenance of the retrieval corpus, not merely on retrieval or generation architecture. \textcite{kotonya2020explainable} operationalise this principle directly in constructing the PUBHEALTH dataset, comprising 11.8 thousand expert-labelled public health claims, and show that veracity prediction models benefit from being paired with journalist-style, evidence-grounded explanations rather than a bare verdict label.

The importance of auditable provenance is made explicit by \textcite{alu2026auditable}, whose conceptual clinical AI decision-support framework ties every generated recommendation to identifiable clinical guidelines through RAG and logs each retrieval and inference step in a tamper-evident audit trail, arguing that source verification and accountability are inseparable from clinical safety in AI-assisted health systems. This principle extends directly to the present study's design rationale: WHO fact sheets provide globally standardised, continuously updated disease information covering transmission, symptoms, prevention, and treatment, while NCDC situation reports provide outbreak-specific, epidemiologically current data disaggregated by state and local government area, including case counts, fatality rates, and affected locations \parencite{who2024factsheets, ncdc2025sitreps}. Neither source alone is sufficient for verifying claims of the kind identified in Section 1: WHO fact sheets establish whether a claim about disease transmission or treatment is generally true, whereas NCDC situation reports establish whether a claim about a specific outbreak, location, or case count is currently accurate. The framework proposed in this study is therefore designed to retrieve from both sources jointly rather than treating them as substitutable.

\subsection{Related Empirical Studies}

Two further empirical studies inform the design and evaluation choices made in this study. \textcite{cao2027medguard} introduce MedGUARD, a two-stage foundation-model collaborative framework that generates multi-dimensional, adversarially negotiated evidence and fuses it with claim features for explainable, hallucination-reduced medical misinformation detection, reporting F1 improvements of 3.2 to 4.5 points over existing benchmarks; the adversarial evidence-negotiation step is a design consideration this study's evaluation protocol draws on when interpreting disagreements between retrieved WHO and NCDC evidence. \textcite{sun2024scoping} conduct a scoping review of natural language processing methods for addressing medically inaccurate information, synthesising the diverse landscape of tasks, datasets, models, and evaluation metrics used across the field and identifying the lack of standardised evaluation as a persistent barrier to comparing systems, a gap this study addresses in part by reporting standard classification metrics (accuracy, precision, recall, and F1) against a purpose-built, labelled claim dataset.

Taken together, the reviewed literature establishes three points that directly motivate the methodology adopted in this study. First, transformer-based classifiers, particularly when grounded in a trusted external source, outperform earlier machine learning baselines on health misinformation tasks. Second, retrieval-augmented generation reliably improves verification accuracy over non-retrieval approaches, with multi-source retrieval designs offering further gains. Third, no existing system combines a global authoritative health source (WHO) with a national, outbreak-specific authority (NCDC) for verifying claims situated in a Nigerian context, the gap this study's proposed framework is designed to close.

\section{Methodology}

\subsection{Research Design}
This study adopts a Design Science Research (DSR) methodology combined with an experimental research design to develop and evaluate a Retrieval-Augmented Transformer (RAT) framework for health misinformation verification. Design Science Research is appropriate because the study focuses on designing, implementing, and evaluating an intelligent artifact capable of verifying health-related claims using evidence retrieved from authoritative public health knowledge sources. The experimental component assesses the effectiveness of the proposed framework by comparing its verification performance using standard classification and retrieval evaluation metrics.

The framework integrates information retrieval techniques with transformer-based language models to retrieve relevant evidence from trusted health repositories before determining the veracity of a health claim. The methodology consists of knowledge source selection, knowledge base construction, health claim dataset preparation, retrieval-augmented verification, and performance evaluation.

\subsection{Knowledge Source Selection}
The reliability of a Retrieval-Augmented Transformer largely depends on the credibility of its external knowledge sources. This study therefore employs official publications from the World Health Organization (WHO) and the Nigeria Centre for Disease Control and Prevention (NCDC) as the primary evidence repositories.

The scope of the knowledge base is limited to five infectious diseases that constitute significant public health concerns in Nigeria and globally, namely Lassa fever, cholera, measles, mpox, and COVID-19. For the WHO repository, official WHO Fact Sheets are selected because they provide standardized and globally accepted information on disease transmission, symptoms, diagnosis, treatment, prevention, and vaccination. For the NCDC repository, official Situation Reports are selected because they contain localized epidemiological information, including outbreak status, affected states, confirmed cases, suspected cases, deaths, surveillance updates, and public health interventions.
Only publicly available documents published by the WHO and NCDC are included to ensure that all retrieved evidence originates from authoritative and verifiable sources.

\subsection{Knowledge Base Construction}
This serves as the external memory of the proposed Retrieval-Augmented Transformer framework. Its purpose is to provide reliable evidence that supports the verification of health-related claims while minimizing hallucinations and misinformation generated by large language models. The knowledge base is constructed through repository creation, document preprocessing, semantic chunking, embedding generation, and vector indexing.

\subsubsection{WHO \& NCDC Knowledge Repository}

The knowledge repository is constructed by integrating official publications from both the World Health Organization (WHO) and the Nigeria Centre for Disease Control and Prevention (NCDC). It contains authoritative information on the five selected diseases: Lassa fever, cholera, measles, mpox, and COVID-19. These documents provide reliable evidence on disease epidemiology, causes, transmission, symptoms, diagnosis, treatment, prevention, vaccination, and public health recommendations from both global and national perspectives.

Official WHO fact sheets and NCDC guidance documents, situation reports, and public health advisories are collected from their respective official websites and converted into machine-readable text. During preprocessing, each document is cleaned while preserving essential metadata, including the disease name, publication date, claim, source, verdict, location, and source URL. The processed documents are then organized into a unified knowledge repository that serves as the trusted evidence base for the retrieval and verification of health-related claims within the proposed Retrieval-Augmented Generation (RAG) framework.


\subsubsection{Document Cleaning and Chunking}
Document preprocessing is performed to improve the quality and consistency of the knowledge base before indexing. The preprocessing stage includes extracting text from PDF and web documents, removing duplicate content, correcting encoding errors, eliminating page headers, footers, page numbers, tables of contents, hyperlinks, and other irrelevant formatting artifacts. Text normalization is also applied to standardize punctuation, spacing, and disease terminology.

After preprocessing, each document is divided into overlapping semantic chunks to facilitate efficient retrieval. Because many health claims reference specific states, dates, outbreak periods, and epidemiological statistics, particularly within the NCDC situation reports, chunk boundaries are carefully designed to preserve these contextual relationships. Chunk overlap is employed to maintain semantic continuity between adjacent segments and reduce the likelihood of separating related information during retrieval.
\subsubsection{Embedding Generation}
However, after document chunking, each text segment is converted into a dense semantic embedding using a pretrained sentence embedding model. Dense embeddings capture the semantic meaning of document content and enable similarity-based retrieval instead of traditional keyword matching.

Each generated embedding is stored together with its corresponding document chunk and metadata, including disease name, publication date, document source, and document identifier. These embeddings provide the semantic representation required for efficient retrieval of supporting evidence.

\subsubsection{Vector Database Construction}

The processed WHO and NCDC documents are converted into dense semantic vector representations to enable efficient retrieval of relevant evidence during claim verification. Each document is divided into smaller, semantically coherent text chunks to improve retrieval granularity and ensure that the system retrieves the most relevant portions of a document instead of an entire report.

Each text chunk is encoded into a high-dimensional embedding using a pre-trained sentence embedding model. These embeddings capture the semantic meaning of the text, enabling the retrieval system to identify relevant evidence even when the wording of a health claim differs from that of the source documents.

The generated embeddings, together with their associated metadata including disease name, document source (WHO or NCDC), publication date, location, source URL, and document identifier—are stored in a vector database. During claim verification, an input health claim is encoded using the same embedding model and compared with the stored embeddings through semantic similarity search. The vector database retrieves the top-k most relevant evidence chunks, which are then provided to the transformer-based classification model for veracity prediction.

By retrieving evidence from authoritative WHO and NCDC sources before classification, the proposed framework grounds its predictions in verifiable information rather than relying solely on the transformer's internal knowledge. This approach improves the transparency, explainability, and reliability of health misinformation verification while reducing the likelihood of unsupported or hallucinated predictions.


\subsection{Health Claim Dataset}

The health claim dataset provides the benchmark for training and evaluating the proposed Retrieval-Augmented Transformer framework. The dataset consists of health-related claims covering the five selected diseases and includes claims that are true, false, or misleading. Each claim is paired with supporting evidence retrieved from the WHO and NCDC repositories.

\subsubsection{Claim Collection}

The study utilizes a curated dataset consisting of 67 health-related claims collected from public sources, including social media posts, news reports, public health communications, and fact-checking platforms. The claims cover Lassa fever, cholera, measles, mpox, and COVID-19.

Several claims contain references to Nigerian states, outbreak dates, confirmed case counts, mortality figures, and other epidemiological information. These contextual details are preserved throughout the retrieval process to enable accurate evidence matching with the corresponding WHO fact sheets and NCDC situation reports.

\subsubsection{Data Cleaning and Annotation}
The collected claims undergo preprocessing to remove duplicate entries, correct formatting inconsistencies, and eliminate incomplete or ambiguous statements. Text normalization is applied to improve consistency while preserving the original meaning of each claim.

Each claim is manually annotated using evidence obtained from the WHO and NCDC repositories and assigned one of three verification labels: True, False, or Misleading. Annotation guidelines are developed to ensure consistency throughout the labeling process. The final annotated dataset is then partitioned into training, validation, and testing subsets for experimental evaluation.


\subsection{Retrieval-Augmented Verification Framework}

\subsubsection{Retriever Module}
The knowledge base's retriever module retrieves evidence for a candidate claim from the knowledge base. The retriever in this implementation uses the sentence-transformer \texttt{all-MiniLM-L6-v2} to encode the candidate claim and subsequently performs a cosine-similarity search against either the chunk-level knowledge base built in Section 3.2 or, when no proper knowledge base is available, a small fallback corpus of five disease-summary sentences (one per disease in scope). For every query, we return the top-$k=3$ most similar fragments or summaries.

\subsubsection{Transformer Verification Module}

The transformer verification module sends and combines the evidence text and the candidate claim, in the form \texttt{EVIDENCE} retrieved\_text [CLAIM] claim\_text, and outputs a probability distribution over the verdict class. The tokeniser applied is the one to be used by the underlying transformer backbone. We set a maximum sequence length of 192 tokens for the evidence and the claim. 

\subsubsection{Evidence Ranking}

The step of ranking the evidence is implicitly applied during the retriever's top-$k$ selection. This selection is done based on cosine similarity and produces a ranking order for all retrieved passages down to the top k passage. The implementation does not apply a re-ranking model, and instead, the ranking of the embedding-based retriever is used in verification.

\subsubsection{Final Claim Classification}

The argmax of the transformer's softmax output over the three classes (True, False, Misleading) produces the final claim classification. The reported metrics are calculated using the human-assigned label of each test claim.

\subsection{Model Fine-Tuning}

We used BERT-base, RoBERTa-base, and DeBERTa-v3-base as candidate classifiers for fine-tuning three transformer encoders.  The training split has 42 samples, each model was fine-tuned on 42-sample training split for up to 10 epochs using Hugging Face \texttt{Trainer} API. The learning rates, batch sizes, and warm-up ratios are in accordance with the current recipe for the architecture, which is $2 \times 10^{-5}$ for BERT and RoBERTa with batch size 16–-32, and $1 \times 10^{-5}$ for DeBERTa-v3 with batch size 8 with 10\% warm-up ratio with cosine learning-rate schedule. The chosen checkpoint was the one that gave the highest validation F1 score. DeBERTa-v3 was further optimised through (i) a class-weighted cross-entropy loss using inverse-frequency class weights and (ii) a label-preserving text-augmentation procedure that generates two paraphrastic variants of each minority-class (True, Misleading) training claim by synonym substitution of disease names and by random stop-word drop.

\subsection{Experimental Setup}

All of the experiments utilized the Hugging Face Transformers library, PyTorch, scikit-learn, and FAISS in Python. The retrieval module and embedding generation uses sentence-transformers library. The training runs were performed in a single Google Colab session. Random seeds were retained for reproducibility.

\subsection{Evaluation Metrics}

The held-out 14-sample test split is used to test the framework on accuracy, weighted precision, weighted recall, and weighted F1 score. Weighted averaging is said to enable comparison with previously published work. However, severe class imbalance limits its use, which we discuss in Section 4 and revisit in Section 5. We make explicit the recommendation that macro-F1 and per-class recall be reported.

\section{Results and Discussion}

\begin{figure}[h!]
    \centering
    \includegraphics[width=0.85\linewidth]{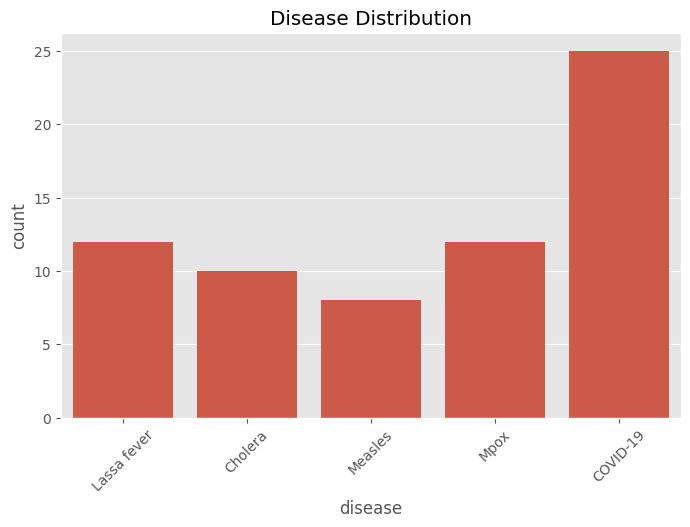}
    \caption{Disease distribution of the 67 health misinformation claims in the dataset.}
    \label{fig:disease-distribution}

    \vspace{0.4cm}

    \includegraphics[width=0.85\linewidth]{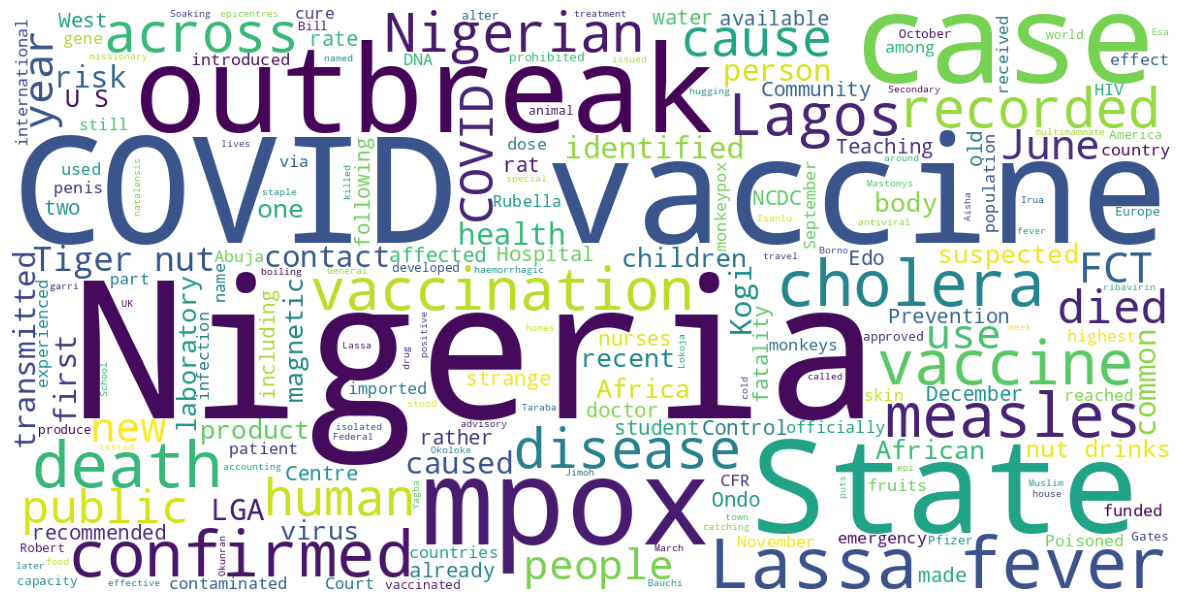}
    \caption{Word cloud of the most frequent terms in the health misinformation dataset.}
    \label{fig:claim-wordcloud}
\end{figure}

\subsection{Dataset Characteristics}

The health misinformation dataset contained 67 fact-checked claims collected from Nigerian and regional fact-checking sources (Dubawa, Africa Check, AFP Fact Check, and NCDC advisories) across five disease categories as shown in Figure~\ref{fig:disease-distribution}. Of these claims, COVID-19 made up the largest proportion (37\%, $n = 25$). The conditions of cholera, mpox, measles and Lassa fever accounted for the remaining 42 claims. Forty two samples useful for training, eleven for validation and fourteen for testing. In the DeBERTa-v3 experiments, class weighting and text augmentation were also applied to counteract class imbalance, which increased the overall training set from 42 to 86 samples.

Figure~\ref{fig:claim-wordcloud} depicts common terms in health misinformation dataset. Key terminologies that dominated claims include Nigeria,\textit{COVID-19}, \textit{case}, \textit{vaccination}, \textit{cholera}, \textit{mpox}, \textit{Lassa fever}, \textit{outbreak}, \textit{disease} etc. this observation shows that these claims which were gathered are strongly centred around surveillance of infectious diseases, communication, vaccination, outbreak reporting etc. It is also noted that terms which are geopolitically relevant such as \textit{Nigeria}, \textit{Lagos} and \textit{African} are often repeated showing a public health context. The word cloud reveals that the corpus captures the key themes of health misinformation in the selected disease domains and provides relevant linguistic content for training and evaluating transformer-based classification models.

\subsection{Performance Evaluation}

Table~\ref{tab:results} presents the test-set results of BERT-base, RoBERTa-base, the baseline DeBERTa-v3, an optimised DeBERTa-v3 (class weighting + text augmentation) and a retrieval-augmented DeBERTa. The BERT-base model achieved the best overall test performance, yielding an overall accuracy of 0.71 and weighted F1 of 0.66. The first model using RoBERTa-base achieved 0.50 accuracy and had a weighted F1-score of 0.33 while the baseline DeBERTa-v3 achieved 0.43 accuracy and had a weighted F1-score of 0.30.

The number of samples in the DeBERTa training set increased from 42 to 86 due to class weighting and text augmentation.  Nevertheless, the DeBERTa model, after being optimised, exhibited an accuracy of 0.36 with a weighted F1-score of 0.19 as demonstrated in the confusion matrix, which placed every test claim in the majority class. This behaviour is characteristic of a single-class prediction bias. Thus, the augmentation strategy directed the model towards the augmented majority class rather than improving discrimination among True, False, and Misleading claims.

The retrieval-augmented DeBERTa model achieved the same accuracy and weighted F1 score as optimized non-retrieval model. In this experiment, retrieval augmentation did not measurably improve performance.  The reason for this finding is the use of a small knowledge corpus with only five general disease summaries as fallback, passage retrieved is not relevant to the submitted claim in some cases.

\begin{table}[h]
\centering
\caption{Test-set performance across model configurations.}
\label{tab:results}
\begin{tabular}{|l|c|c|c|c|}
\hline
Model & Accuracy & Precision & Recall & F1 \\
\hline
\textbf{BERT-base}                        & \textbf{0.71} & \textbf{0.61} & \textbf{0.71} & \textbf{0.66} \\
\hline
RoBERTa-base                      & 0.50 & 0.25 & 0.50 & 0.33 \\
\hline
DeBERTa-v3-base                   & 0.43 & 0.23 & 0.43 & 0.30 \\
\hline
DeBERTa + class weighting + augmentation & 0.36 & 0.13 & 0.36 & 0.19 \\
\hline
DeBERTa + RAG (evidence-augmented)       & 0.36 & 0.13 & 0.36 & 0.19 \\
\hline
\end{tabular}
\end{table}

\subsection{Comparison with Baseline Models}

The arrangement of three backbones on the test split is BERT-base $\succ$ better than RoBERTa-base $\succ$ better than DeBERTa-v3-base. The above ordering is consistent
with the validation results of the same models. The validation accuracies of BERT-base and RoBERTa-base are 0.55 each. The validation accuracy of DeBERTa-v3-base is 0.50. A small performance gap on the validation and test set for the better-regularised backbones (BERT and RoBERTa) indicates that these models generalise reasonably well from the very small training set. However, DeBERTa-v3, which is designed to be sample-efficient on larger training corpora, seems to be more sensitive to the extreme class imbalance and to the lack of strong inductive bias in the training data.

It is notable that the optimisation interventions (class weighting and text augmentation) do not help improve DeBERTa-v3 test performance. By doubling the size of training sets for True and Misleading class, author proposes an augmentation procedure. However, the additional samples are paraphrase on surface-level of the original claim but are essentially the same claim, thus model keeps predicting the augmented majority class for every test instance. This paper’s finiding is consistent with the large literature on fine-tuning on small data. In general, oversampling or augmenting minority classes can shift the decision boundary in the desired direction, but it cannot replace a sufficient number of independently sourced training examples. 

\subsection{Error Analysis}

The error analysis of the DeBERTa-v3 model on the test set indicates a single-class prediction pattern, in which the model predicts the majority class for the entire test split. Both the validation and test confusion matrices show non-zero entries only on one row, indicating that after a few fine-tuning epochs, the model collapsed to the majority class. Softmax output of the model gives high confidence values for the class predicted, meaning that it is confident that it is wrong rather than uncertain. This behaviour is typical of fine-tuning a moderately large transformer on a very tiny and imbalanced dataset, where the cross-entropy loss is dominated by the majority class and the model converges to a degenerate solution.

In the present setup, the retrieval evidence did not appear to affect the model’s predictions as the retrieval-augmented configuration produced an identical confusion matrix to the optimised nonretrieval configuration. An illustrative case in point highlighted the DII that was extracted from FDI. The DII’s feasibility is constrained by factors including (i) the absence of a collection of research reports associated with the measure; (ii) the unavailability of both standalone and collective AICD; and (iii) the lack of universally accepted ICD.

\subsection{Discussion}

The results of this analysis bring to light two findings likely to be of interest to the wider community working on automated verification of health misinformation in low-resource settings. Initially, it was discovered that the ranking of off-the-shelf transformer backbones, on a 42-sample training dataset from the Nigerian information environment, is not the one observed on large-benchmarks. The easiest of the three architectures, BERT-base delivered the best performance on the test set, and the more recent sample-efficient DeBERTa-v3 base model does not outperform either of the older models. The present data scale indicates that, the inductive bias of the model and the pretraining objective matter less than the fine-tuning protocol’s ability to find stable solutions in a heavily imbalanced training set.

Secondly, the retrieval-augmented version of DeBERTa-v3 has no measurable improvement compared to DeBERTa-v3,          thus the literature has an exaggerated value of retrieval augmentation for knowledge-intensive verification tasks. The most reasonable explanation, borne out in the review of passages retrieved, is that the fallback knowledge corpus used in the current experiment is too small and too generic to provide the disambiguating evidence for separating True, False and Misleading claims in the Nigerian information environment. The main direction for future work listed in Section 5 is replacing this backup with an indexed document and section level corpus of WHO fact sheets and NCDC situation reports.


\section{Limitations and Future Work}
Overall, the experiment demonstrates the feasibility of combining semantic retrieval with transformer classification. But data scarcity, a class-imbalanced dataset, single-class model behavior, and a limited evidence corpus constrained the framework's performance. Given the important limitations of this study, future works are listed in bullet points.
\begin{enumerate}
    \item Replace the fallback summaries with a structured knowledge base developed directly from WHO and NCDC documents. Each retrieved passage should preserve its organization, title, publication date, URL, disease category, and unique identifier.
    \item The transformer should be trained and evaluated using evidence--claim pairs so that it learns whether retrieved evidence supports, contradicts, or is insufficient to verify a claim.
    \item The dataset should be expanded and balanced, particularly for the misleading class, while also adopting class-imbalance-aware evaluation (macro-F1, per-class
    recall) as the primary reported metric, since weighted accuracy is
    misleading under severe collapse.
   
\end{enumerate}

\section{Conclusion}
This study compared BERT, RoBERTa, and DeBERTa-v3 for classifying
Nigerian health misinformation claims, and a retrieval-augmented
variant of DeBERTa-v3 using WHO-informed evidence context. BERT-base
achieved the test performance (71\% accuracy, F1 = 0.66); RoBERTa,
DeBERTa-v3, and the optimized/RAG variants all fall biased to single-class
prediction, underscoring that model architecture and retrieval
augmentation could not compensate for an insufficiently sized training set
(n = 42).

\printbibliography

\end{document}